\documentclass[a4paper]{cas-sc}

\usepackage[numbers]{natbib}

\usepackage{amsmath}
\usepackage{amsfonts}
\usepackage{amssymb}
\usepackage{bm}

\DeclareMathOperator{\Format}{Format}
\DeclareMathOperator{\GroupRisk}{GroupRisk}
\DeclareMathOperator{\ExtractLabel}{ExtractLabel}
\DeclareMathOperator{\fields}{fields}
\makeatletter
\g@addto@macro\normalsize{%
    \setlength{\abovedisplayskip}{7pt plus 2pt minus 2pt}%
    \setlength{\belowdisplayskip}{7pt plus 2pt minus 2pt}%
    \setlength{\abovedisplayshortskip}{4pt plus 2pt minus 1pt}%
    \setlength{\belowdisplayshortskip}{5pt plus 2pt minus 1pt}%
}
\makeatother
\newcommand{\riskSupport}{\mathrm{support}}
\newcommand{\riskAgainst}{\mathrm{against}}
\newcommand{\riskUncertain}{\mathrm{uncertain}}

\usepackage{graphicx}
\usepackage{xcolor}
\usepackage{tikz}
\usepackage{transparent}

\usepackage{placeins}
\usepackage{booktabs}
\usepackage{array}
\usepackage{tabularx}
\usepackage{threeparttable}
\usepackage{colortbl}
\usepackage{ragged2e}
\usepackage{siunitx}
\usepackage{multirow}
\usepackage{makecell}

\newcolumntype{P}[1]{>{\RaggedRight\arraybackslash}p{#1}}
\newcolumntype{Z}[1]{>{\Centering\arraybackslash}p{#1}}
\newcolumntype{Y}{>{\RaggedRight\arraybackslash}X}
\definecolor{mygray}{gray}{0.9}
\definecolor{myGreen}{RGB}{169, 217, 187}
\definecolor{myBlue}{RGB}{144, 180, 207}
\definecolor{tableBlue}{RGB}{146, 184, 210}
\definecolor{tableGreen}{RGB}{173, 220, 193}
\definecolor{tableStripe}{RGB}{244, 244, 244}
\definecolor{tableRule}{RGB}{20, 20, 20}

\newcommand{\TableBand}[3]{%
    \rowcolor{#1}\multicolumn{#2}{c}{\rule{0pt}{2.15ex}\small\bfseries #3}\\%
}
\newcommand{\TableBandMain}[3]{%
    \rowcolor{#1}\multicolumn{#2}{c}{\rule{0pt}{2.35ex}\normalsize\bfseries #3}\\%
}

\usepackage{algorithm}
\usepackage{algpseudocode}

\algrenewcommand\algorithmiccomment[1]{\hfill{\footnotesize$\triangleright$ #1}}
\newcommand{\AlgStage}[1]{%
    \Statex\vspace{0.25ex}%
    \Statex{\bfseries\itshape #1}%
    \Statex\vspace{0.1ex}%
}

\usepackage{pifont}

\usepackage{orcidlink}
\usepackage{etoolbox}
\usepackage{xspace}

\def\tsc#1{\csdef{#1}{\textsc{\lowercase{#1}}\xspace}}
\tsc{WGM}
\tsc{QE}

\usepackage[switch]{lineno}

\usepackage{hyperref}
\hypersetup{
    colorlinks=true,
    linkcolor=cyan,
    filecolor=cyan,
    urlcolor=cyan,
    citecolor=cyan
}

\def\tsc#1{\csdef{#1}{\textsc{\lowercase{#1}}\xspace}}
\tsc{WGM}
\tsc{QE}

\begin{document}
\let\WriteBookmarks\relax
\let\printorcid\relax

\renewcommand{\topfraction}{0.95}
\renewcommand{\bottomfraction}{0.95}
\renewcommand{\textfraction}{0.05}
\renewcommand{\floatpagefraction}{0.85}
\setcounter{topnumber}{5}
\setcounter{bottomnumber}{5}
\setcounter{totalnumber}{10}

\def\floatpagepagefraction{1}
\def\textpagefraction{.001}
\let\printorcid\relax 

\shorttitle{} 

\shortauthors{  \textit{et al.}}

\title[mode = title]{Structured Visual Evidence Decomposition for Evidence-Grounded Multimodal Screening of Obstructive Sleep Apnea-Hypopnea Syndrome}

\author[1]{Chen Zhan\orcidlink{0009-0003-1911-0242}}

\author[1]{Yingchen	Wei}

\author[2]{Xiaoyu Tan\orcidlink{0000-0003-3555-7143}}

\author[3]{Jingjing	Huang}

\author[1,4]{Xihe Qiu\orcidlink{0000-0003-4024-925X}}
\cormark[1]

\address[1]{School of Electronic and Electrical Engineering, Shanghai University of Engineering Science, Shanghai 200438, China}
\address[2]{Tencent Youtu Lab, Shanghai 200233, China}

\address[3]{ENT Institute and Department of Otorhinolaryngology, Eye \& ENT Hospital of Fudan University, Shanghai 200031, China}
\address[4]{National University of Singapore, 9 Engineering Drive 1, Singapore 117575, Singapore}

\cortext[1]{Corresponding author.}

\begin{abstract}
Effective pre-polysomnography screening for Obstructive Sleep Apnea-Hypopnea Syndrome (OSAHS) requires integrating structured clinical risk factors with visible craniofacial and neck morphological cues. However, directly prompting general-purpose multimodal foundation models to issue medical yes/no decisions can produce unstable and poorly calibrated outputs, including degenerate negative predictions or excessive positive calls. We propose EviOSAHS, an evidence-grounded multimodal reasoning framework that separates image-only anatomical evidence acquisition from final clinical adjudication. The framework first decomposes each frontal facial image into seven fixed anatomical queries targeting the neck, chin, mouth, face/neck fat, lower jaw, midface, and nose. Each visual response is then converted into a structured evidence card containing the anatomical target, visibility, risk direction, evidence strength, confidence, and a concise evidence summary. Finally, the evidence cards are integrated with a cleaned structured clinical profile only at the final stage, where a large language model performs balanced binary screening adjudication. We evaluated EviOSAHS on a 642-subject cohort for binary OSAHS screening, with normal subjects mapped to screening-negative and all mild, moderate, or severe OSAHS subjects mapped to screening-positive. EviOSAHS achieved 88.47\% accuracy, 94.86\% sensitivity, 93.74\% F1-score, and a 5.14\% false-negative rate. Under a unified protocol, EviOSAHS outperformed clinical-only prompting, direct multimodal prompting, and naive two-stage pipelines, with paired McNemar tests confirming sample-level advantages. Ablation studies confirmed that seven-question visual decomposition and balanced final adjudication were critical to the high-sensitivity operating point. A question-level audit of 4,494 visual outputs yielded a 100\% structured parse rate and a 93.88\% high-visibility rate, supporting the formatting reliability of image-only evidence acquisition. Error analysis showed that the operating point prioritized missed-case reduction but required better calibration when multiple weak risk cues accumulated. Structured visual evidence decomposition can organize general-purpose multimodal foundation models into an auditable, high-sensitivity workflow for binary pre-polysomnography OSAHS screening. The proposed framework should be interpreted as a pre-polysomnography triage assistant rather than a definitive diagnostic system; prospective validation, external testing, and calibrated operating-point control are required before clinical deployment.

\end{abstract}

\begin{highlights}
    \item We formulate OSAHS assessment as high-sensitivity pre-polysomnography screening from facial images and structured clinical profiles.
    \item The framework separates image-only anatomical evidence acquisition from final clinical adjudication.
    \item Seven anatomy-specific visual queries convert facial observations into auditable evidence cards.
    \item On 642 subjects, the proposed method achieved 88.47\% accuracy, 94.86\% sensitivity, 93.74\% F1-score, and 5.14\% FNR.
    \item Paired tests, ablations, visual-output audits, image controls, and error attribution characterize missed-case reduction and remaining calibration limitations.
\end{highlights}


\maketitle

\section{Introduction}

Obstructive Sleep Apnea-Hypopnea Syndrome (OSAHS) is a common and clinically consequential sleep-related breathing disorder associated with hypertension, cardiovascular disease, neurocognitive impairment, daytime dysfunction, and reduced quality of life~\cite{benjafield2019global,gottlieb2020jama,yeghiazarians2021aha,javaheri2017jacc,salman2020currcardiol}. Although patients may present with habitual snoring, witnessed apnea, fragmented sleep, frequent nocturnal awakening, and excessive daytime sleepiness~\cite{chung2016stopbang,laratta2017cmaj}, symptom-based assessment alone does not fully capture individual screening risk. OSAHS susceptibility is also influenced by upper-airway and craniofacial phenotypes, including mandibular retrusion, crowded oral structures, soft-tissue accumulation, and enlarged neck morphology~\cite{vicini2012nohl,agha2017facialphenotype}. These complementary signals indicate that effective OSAHS screening should jointly consider structured clinical risk factors and visible craniofacial or neck morphological cues rather than relying on either modality alone.

\begin{figure}[pos=htbp]
\centering
\includegraphics[width=0.90\textwidth]{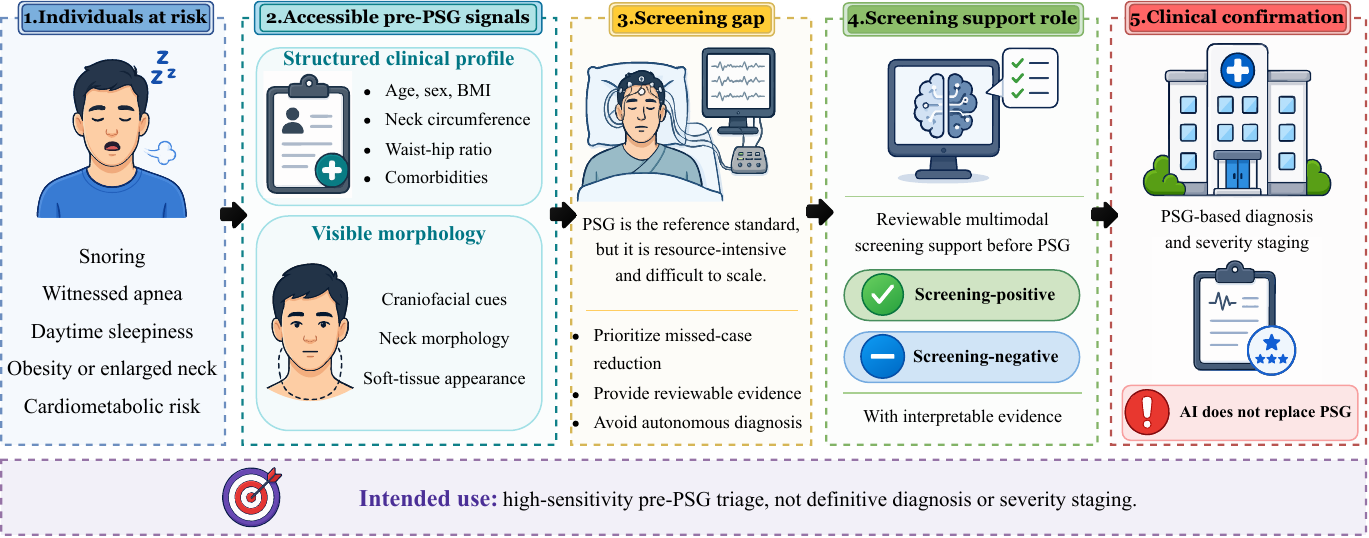}
\caption{\textbf{Clinical motivation and intended use of multimodal pre-polysomnography OSAHS screening.} Patients at risk for OSAHS may present with symptoms, structured clinical risk factors, and visible craniofacial or neck morphological cues. Because PSG remains the reference standard for diagnosis and severity staging but is resource-intensive and difficult to scale for broad preliminary assessment, a pre-PSG screening tool should prioritize missed-case reduction while providing evidence that can be reviewed by clinicians. The intended use is high-sensitivity triage before confirmatory PSG, not autonomous diagnosis or severity grading.}
\label{fig1}
\end{figure}

Overnight polysomnography (PSG) remains the reference standard for OSAHS diagnosis and severity staging, with the apnea-hypopnea index (AHI) commonly used to define clinical severity categories~\cite{kapur2017aasm,malhotra2021beyondahi}. However, PSG is resource-intensive, time-consuming, and difficult to scale for all individuals who require preliminary risk assessment. A practical pre-PSG screening tool should therefore be optimized for missed-case reduction, provide reviewable evidence for clinician oversight, and avoid using label-derived information such as AHI during prediction. Accordingly, the objective of this study is not to replace PSG, infer disease severity, or deliver autonomous diagnosis, but to support high-sensitivity triage before confirmatory testing. Figure~\ref{fig1} summarizes this clinical context and defines the intended use of multimodal pre-PSG screening as clinician-reviewable triage support.


Previous computational approaches have explored OSAHS risk assessment using structured clinical profiles, facial images, acoustic signals, oximetry, video analysis, and other accessible modalities ~\cite{ge2024osahsml}. Classical machine learning and deep learning models can learn discriminative patterns from these data sources, but many existing systems either require task-specific training or formulate multimodal fusion as a black-box classification problem~\cite{park2025craniofacial,monna2022craniofacial3d,kim2023cephalogram}. This limits clinical interpretability and makes it difficult to determine whether a prediction is driven by meaningful evidence or by spurious correlations. Unimodal systems are also inherently constrained: a text-only model may overlook visible craniofacial or neck cues, whereas an image-only model cannot adequately account for age, sex, body habitus, comorbidities, and other structured clinical risk factors~\cite{abad2016video,jiang2021snoring,shen2020snore}. Therefore, a clinically useful screening workflow should integrate visual and clinical information while preserving a transparent record of how each source contributes to the final decision.

Recent advances in large language models (LLMs) and vision-language models (VLMs) provide new opportunities for multimodal clinical decision support. VLMs can translate visual content into natural-language observations~\cite{radford2021clip,li2023blip2,dai2023instructblip,liu2023llava,bai2025qwen25vl}, whereas LLMs can reason over structured and unstructured clinical information~\cite{yang2024qwen25}. A straightforward strategy is to place the patient image and clinical summary into a single multimodal prompt and directly request an OSAHS yes/no judgment~\cite{singhal2023clinical,thirunavukarasu2023llmmedicine}. However, direct multimodal prompting is not a reliable substitute for a clinically controlled screening workflow. Its limitations can be understood from three aspects: visual evidence acquisition is unconstrained, visual and clinical information are entangled too early, and the final decision is generated without an explicit balance of supporting, opposing, and uncertain evidence. These limitations make the decision boundary difficult to audit or calibrate~\cite{liu2024lvlmhallucination,kapoor2024knowunknown}. Consistent with these concerns, our experiments showed that direct prompting either failed to outperform a clean clinical-only baseline or collapsed toward degenerate decision patterns. Therefore, the central challenge is not merely whether images and clinical text can be placed in the same prompt, but whether multimodal foundation models can be organized to acquire, structure, and adjudicate evidence in a clinically controlled manner.

To address this challenge, we propose EviOSAHS, an evidence-grounded multimodal reasoning framework for high-sensitivity OSAHS screening. EviOSAHS converts general-purpose foundation models from direct clinical classifiers into a staged evidence-acquisition and adjudication workflow. A frontal facial image is decomposed into seven fixed anatomical questions targeting the neck, chin, mouth, face-and-neck fat accumulation, lower jaw, midface, and nose. At this stage, the VLM receives only the image and is constrained to act as an anatomy-specific visual observer rather than a diagnostic classifier. The resulting visual observations are converted into structured evidence cards that record the anatomical target, visibility, risk direction, evidence strength, confidence, and a concise evidence summary. These evidence cards are integrated with a cleaned structured clinical profile only at the final decision stage, where an LLM performs balanced binary screening adjudication.

This design directly addresses the limitations of direct prompting. Visual perception and clinical reasoning are decoupled so that intermediate image-derived evidence can be inspected and audited. Clinical information is introduced only after visual evidence acquisition, reducing the risk that clinical priors overwrite image-grounded observations. The final decision is based on explicit evidence aggregation rather than unconstrained generation. In this study, balanced adjudication refers to a constrained final decision procedure that compares supporting, opposing, and uncertain evidence before integrating structured clinical risk factors. By converting visual descriptions into evidence cards and requiring balanced adjudication, EviOSAHS enables targeted ablation, error analysis, and clinician-facing interpretation.

The main contributions of this study are fourfold:
\begin{enumerate}
    \item We formulate OSAHS assessment as a high-sensitivity binary pre-polysomnography screening task from frontal facial images and structured clinical profiles, explicitly separating screening-positive identification from autonomous diagnosis and exploratory severity grading.
    \item We propose EviOSAHS, an evidence-grounded multimodal reasoning framework that constrains the VLM to image-only anatomical observation and converts visual responses into structured, auditable evidence cards.
    \item We introduce a final-only clinical adjudication strategy that delays structured clinical information until the decision stage, reducing premature clinical-prior contamination of image-derived evidence.
    \item We provide a unified evaluation on a 642-subject cohort against clinical-only, direct multimodal, naive two-stage, early clinical-fusion, single-model, and backbone-transfer variants, supported by paired testing, ablation, visual-output auditing, image controls, subgroup analysis, and error attribution.
\end{enumerate}


\section{Related Work}

This study builds on three research directions: OSAHS screening and craniofacial risk modeling, multimodal foundation models for clinical AI, and structured reasoning for evidence-grounded adjudication. Rather than reviewing each area exhaustively, we focus on the remaining gaps that motivate structured visual evidence decomposition for pre-PSG screening.

\subsection{OSAHS screening and craniofacial risk modeling}

OSAHS diagnosis is anchored by overnight polysomnography, yet the need for scalable preliminary risk stratification has motivated screening approaches based on accessible clinical, anatomical, and physiological signals~\cite{kapur2017aasm,malhotra2021beyondahi}. Clinical questionnaires and risk scores, such as STOP-Bang, use symptoms, BMI, age, sex, neck circumference, and comorbidities to identify individuals at elevated risk~\cite{chung2016stopbang}. Machine learning studies have further shown that structured demographic and cardiometabolic variables can support OSAHS risk prediction~\cite{ge2024osahsml}. However, clinical-only screening cannot directly inspect visible craniofacial and neck morphology, which may reflect anatomical susceptibility to upper-airway obstruction~\cite{vicini2012nohl,agha2017facialphenotype}.

Image- and morphology-based studies address this limitation by using facial photographs, 3D craniofacial scans, or lateral cephalograms for OSAHS risk assessment~\cite{park2025craniofacial,monna2022craniofacial3d,kim2023cephalogram}. Signal- and video-based methods have also explored snoring sounds, sleep video, oximetry, and other physiological measurements~\cite{abad2016video,jiang2021snoring,shen2020snore}. These studies demonstrate that non-PSG signals can provide clinically meaningful screening information. Nevertheless, many existing systems rely on task-specific training, specialized acquisition protocols, or direct predictive modeling. They rarely expose an explicit, reviewable record of how visible anatomy contributes to the final screening decision. This gap motivates an auditable image-plus-clinical workflow for pre-PSG triage.

\subsection{Foundation models, medical facial phenotyping, and multimodal clinical AI}

Large-scale VLMs and LLMs have expanded the design space for multimodal clinical AI. VLMs such as CLIP, BLIP-2, InstructBLIP, LLaVA, and Qwen-VL variants can align images with language and generate image-grounded textual responses~\cite{radford2021clip,li2023blip2,dai2023instructblip,liu2023llava,bai2025qwen25vl}. LLMs, in contrast, have shown strong capabilities in medical question answering, clinical text interpretation, and knowledge-intensive reasoning~\cite{singhal2023clinical,thirunavukarasu2023llmmedicine,nori2023gpt4medical}. Medical VLMs such as MedCLIP and BioViL further suggest that image-text alignment can improve domain-specific medical image understanding~\cite{wang2022medclip,boecking2022biovil}.

Facial photographs can also encode clinically meaningful phenotypic information, as shown by computer-aided facial phenotyping systems for genetic and rare disease analysis~\cite{gurovich2019deepgestalt,hsieh2022gestaltmatcher,reiter2024facialtools}. However, OSAHS differs from syndrome-specific facial phenotyping because its visible cues are weaker, less specific, and must be interpreted jointly with structured clinical risk factors. For this reason, general-purpose VLMs should not be treated as direct OSAHS classifiers. A more controlled role assignment is needed: the VLM acquires anatomy-focused visual observations, while the LLM performs clinical interpretation only after those observations have been structured.

\subsection{Structured reasoning and evidence-grounded clinical adjudication}

Structured prompting methods, including chain-of-thought prompting, self-consistency, and ReAct, can improve problem decomposition in LLM workflows~\cite{wei2022cot,wang2023selfconsistency,yao2023react}. However, reasoning structure alone does not ensure clinically grounded decision-making. A fluent reasoning trace may still rely on hallucinated findings, uncontrolled clinical priors, or poorly calibrated visual interpretation~\cite{liu2024lvlmhallucination,kapoor2024knowunknown}. For medical screening, the key requirement is not only to generate intermediate reasoning, but to control what evidence is acquired, when clinical context is introduced, and how conflicting cues are adjudicated.

Our framework therefore differs from generic reasoning prompts and prior direct multimodal classifiers. It treats OSAHS screening as an evidence acquisition and adjudication problem: image-only anatomical queries first produce localized visual observations; these observations are converted into structured evidence cards; and the final LLM adjudicator integrates them with a clean clinical summary only at the decision stage. This explicit separation of evidence acquisition, evidence organization, and clinical adjudication is designed to support auditable high-sensitivity pre-PSG triage rather than autonomous diagnosis.


\section{Methods}

\subsection{Problem formulation}

We formulate OSAHS assessment as a high-sensitivity binary screening task rather than autonomous diagnosis or severity grading. For each subject, the inference inputs consist of a frontal facial image and a structured clinical profile. For task definition and evaluation, each subject is also associated with an AHI-derived reference severity category. The study cohort is denoted as
\begin{equation}
\begin{alignedat}{3}
\mathcal{D}
&=
\{\mathcal{S}_n\}_{n=1}^{N},\qquad&
\mathcal{S}_n
&=
\bigl(\mathcal{I}_n,\mathbf{x}_n,z_n\bigr),\qquad&
z_n
&\in
\mathcal{Z}=\{0,1,2,3\},
\end{alignedat}
\label{eq:subject_tuple}
\end{equation}
where $\mathcal{I}_n$ is the frontal facial image, $\mathbf{x}_n$ is the structured clinical profile, and $z_n \in \{0,1,2,3\}$ is the original AHI-derived severity category corresponding to normal, mild, moderate, and severe OSAHS. The primary endpoint is a binary screening label $y_n \in \{0,1\}$, where normal subjects are screening-negative and all OSAHS severity levels are screening-positive:
\begin{equation}
y_n
=
\mathbb{1}\{z_n>0\}
=
\begin{cases}
0, & z_n = 0,\\
1, & z_n \in \{1,2,3\},
\end{cases}
\label{eq:binary_label_mapping}
\end{equation}

The goal is to generate a screening prediction $\hat{y}_n \in \{0,1\}$ while preserving explicit intermediate evidence. A positive output denotes screening-positive status for pre-polysomnography triage, not definitive OSAHS diagnosis. This distinction determines the design of the workflow: instead of asking a multimodal model to directly diagnose OSAHS from an image and clinical text, the proposed method first acquires localized visual evidence, then organizes this evidence into structured cards, and finally performs clinical adjudication using both the evidence cards and standardized clinical context.

\begin{figure}[pos=htbp]
\centering
\includegraphics[width=0.95\textwidth]{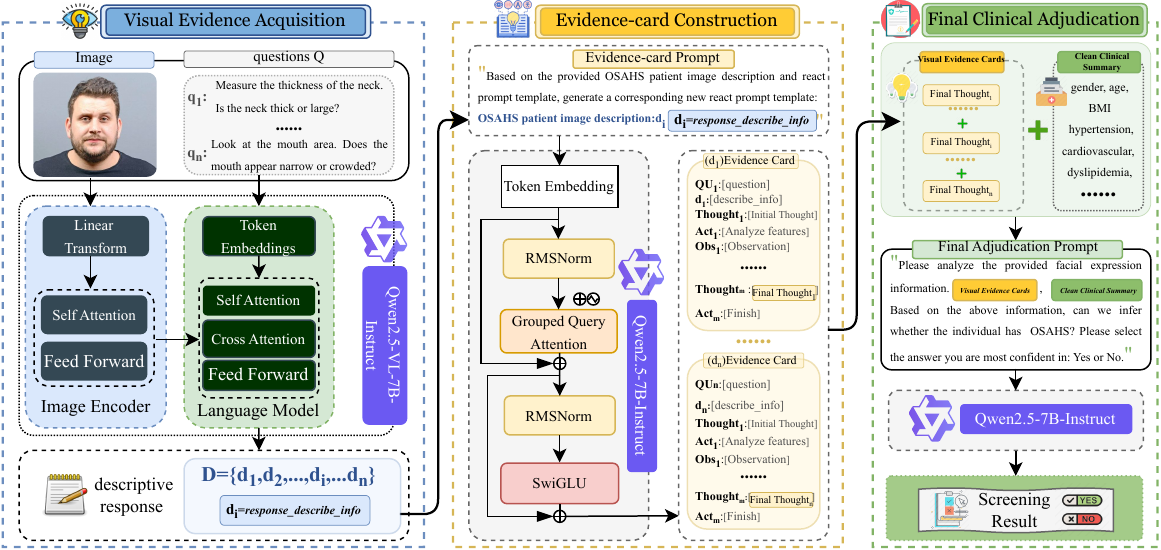}
\caption{\textbf{Overview of the proposed EviOSAHS workflow for evidence-grounded OSAHS screening.}
The framework separates image-only anatomical evidence acquisition from final clinical adjudication. A VLM first extracts localized facial and neck observations through seven fixed anatomical queries. The resulting observations are converted into structured evidence cards containing visibility, risk direction, evidence strength, confidence, and concise summaries. An LLM then integrates the evidence cards with a clean structured clinical summary to generate a binary pre-polysomnography screening result.}
\label{fig:framework}
\end{figure}

\subsection{Framework overview}

The proposed framework, termed \textit{EviOSAHS}, is an evidence-grounded multimodal reasoning workflow for binary pre-PSG OSAHS screening. As shown in Figure~\ref{fig:framework}, EviOSAHS contains four connected components: structured clinical summary reconstruction, anatomy-specific visual evidence acquisition, evidence-card construction, and final-only clinical adjudication. The central design principle is temporal separation between perception and adjudication: visual evidence is acquired from the image alone, while clinical variables are introduced only after visual evidence has been structured.

The overall mapping is written as a staged composition:
\begin{equation}
\label{eq:overall_mapping}
\mathcal{C}_n = g_{\mathrm{clin}}(\mathbf{x}_n), \quad \mathcal{E}_n = G_{\mathrm{vis}} \bigl( \mathcal{I}_n; \mathcal{Q}, T_{\mathrm{vis}}, T_{\mathrm{card}} \bigr), \quad \hat{y}_n = \mathcal{F}(\mathcal{I}_n, \mathbf{x}_n) = f_{\mathrm{final}} \bigl( \mathcal{E}_n, \mathcal{C}_n; T_{\mathrm{final}} \bigr),
\end{equation}
where $\mathcal{C}_n$ denotes the clean clinical summary reconstructed from $\mathbf{x}_n$, and $\mathcal{E}_n$ denotes the ordered set of structured visual evidence cards derived from $\mathcal{I}_n$. Here, $f_{\mathrm{VLM}}$ denotes the image-conditioned visual observer, $f_{\mathrm{card}}$ denotes the LLM-based evidence-card generator, and $f_{\mathrm{final}}$ denotes the LLM-based final adjudicator. The framework does not require the evidence-card generator and final adjudicator to share the same backbone, although the primary implementation in this study uses a Qwen-based instantiation.

This formulation deliberately differs from direct multimodal prompting, where the image and clinical text are placed into a single prompt and the model is asked to issue a yes/no decision. In EviOSAHS, the final answer is delayed until the model has produced an auditable representation of visual evidence. This staged dependency makes the decision pathway inspectable: the final screening output can be traced back to anatomy-specific observations, evidence direction, evidence strength, model-reported confidence, and structured clinical context.

Procedurally, each subject is processed as follows. The structured clinical variables are first converted into a clean clinical summary. The facial image is then queried by a fixed set of anatomy-specific visual questions. Each visual response is parsed into a structured observation and converted into an evidence card. The seven evidence cards are finally combined with the clean clinical summary to generate the binary screening output. The workflow returns both the final label and the evidence-card list, allowing each output to be reviewed not only as a yes/no decision but also as a sequence of anatomical observations and evidence assignments.

This staged design serves two purposes. First, it reduces the risk that clinical priors, such as high BMI or large neck circumference, contaminate what the VLM reports from the image. Second, it converts an otherwise opaque image-text decision into a reviewable evidence acquisition and adjudication process.

\begin{table}[pos=htbp]
\centering
\caption{Anatomical query inventory and structured evidence-card schema used in EviOSAHS. Panel A defines the fixed image-only anatomical questions. Panel B defines the structured fields used to convert each visual observation into an auditable evidence card.}
\label{tab:visual_evidence_schema}
\footnotesize
\setlength{\tabcolsep}{5.5pt}
\renewcommand{\arraystretch}{1.18}
\arrayrulecolor{tableRule}

\begin{tabularx}{\textwidth}{@{} Z{0.8cm} P{3.0cm} Y @{}}
\toprule
\TableBand{tableBlue}{3}{Panel A. Anatomy-specific visual query inventory}
\midrule
\textbf{No.} & \textbf{Target} & \textbf{Visual evidence sought} \\
\midrule
\rowcolor{tableStripe}
1 & Neck & Neck thickness, enlargement, or visually apparent neck soft-tissue accumulation \\
2 & Chin & Chin recession, retrusion, or reduced anterior chin projection \\
\rowcolor{tableStripe}
3 & Mouth & Oral narrowing, crowding, or externally visible cues suggesting limited oral space \\
4 & Face/neck fat & Soft-tissue or fat accumulation around the face and neck \\
\rowcolor{tableStripe}
5 & Lower jaw & Small, retruded, or posteriorly positioned mandible \\
6 & Midface & Midface flattening, underdevelopment, or reduced midfacial projection \\
\rowcolor{tableStripe}
7 & Nose & Visible nasal asymmetry or external cues potentially related to nasal obstruction \\
\bottomrule
\end{tabularx}

\vspace{3pt}

\begin{tabularx}{\textwidth}{@{} P{2.8cm} P{3.2cm} Y @{}}
\toprule
\TableBand{tableGreen}{3}{Panel B. Evidence-card schema}
\midrule
\textbf{Field} & \textbf{Allowed value or type} & \textbf{Role in the workflow} \\
\midrule
\rowcolor{tableStripe}
Anatomical target & One of seven targets & Anchors each observation to a fixed anatomical region \\
Visual observation & Short image-derived text & Records the VLM-generated anatomical observation without clinical variables \\
\rowcolor{tableStripe}
Visibility & high / medium / uncertain & Indicates whether the anatomical region is sufficiently visible for assessment \\
Risk direction & support / against / uncertain & Specifies whether the cue supports, argues against, or is indeterminate for OSAHS screening risk \\
\rowcolor{tableStripe}
Evidence strength & weak / moderate / strong & Qualitatively describes the contribution of the cue to screening relevance \\
Confidence & low / medium / high & Records model-reported certainty; not interpreted as a calibrated probability \\
\rowcolor{tableStripe}
Evidence summary & Concise text & Provides a short reviewable explanation for downstream adjudication \\
\bottomrule
\end{tabularx}
\end{table}

\subsection{Structured clinical summary reconstruction}

The structured clinical profile contains demographic, anthropometric, and comorbidity-related variables. We represent the clinical input as
\begin{equation}
\mathbf{x}_n
=
\bigl[
\mathrm{Age}_n,
\mathrm{Sex}_n,
\mathrm{BMI}_n,
\mathrm{NC}_n,
\mathrm{WHR}_n,
\mathrm{HTN}_n,
\mathrm{DM}_n,
\mathrm{HD}_n,
\mathrm{HLD}_n
\bigr]^{\top},
\label{eq:clinical_vector}
\end{equation}
where $\mathrm{NC}$ denotes neck circumference, $\mathrm{WHR}$ denotes waist-hip ratio, $\mathrm{HTN}$ denotes hypertension, $\mathrm{DM}$ denotes diabetes, $\mathrm{HD}$ denotes heart disease, and $\mathrm{HLD}$ denotes hyperlipidemia. These variables are not passed to the VLM during visual observation; they are used only to construct the clinical context for final adjudication.

A deterministic reconstruction function converts these fields into a standardized clinical summary:
\begin{equation}
\label{eq:clinical_summary}
\mathcal{C}_n = g_{\mathrm{clin}}(\mathbf{x}_n) = \bigl( \mathcal{P}_n, \mathcal{R}^{+}_n, \mathcal{R}^{0}_n \bigr), \quad \operatorname{fields}(\mathcal{C}_n) \cap \{\mathrm{AHI}, z_n, y_n\} = \varnothing,
\end{equation}
where $\mathcal{P}_n$ denotes the patient profile, $\mathcal{R}^{+}_n$ denotes positive clinical risk information, and $\mathcal{R}^{0}_n$ denotes neutral, protective, or unavailable clinical context. The reconstructed summary uses consistent field names and categorical descriptions, such as BMI category and waist-hip ratio category, to reduce prompt instability caused by heterogeneous raw text.

Unavailable fields are explicitly marked as unavailable rather than imputed, so that the final adjudicator can distinguish missing information from normal findings. No AHI value, severity category, or label-derived information is included in $\mathcal{C}_n$, preventing leakage from the reference standard into the screening workflow. Therefore, the clinical summary is not a prediction target and not a shortcut to the ground-truth label. Its role is to provide standardized clinical context after image-derived evidence has already been acquired and organized. This final-only use of $\mathcal{C}_n$ is one of the main differences between EviOSAHS and early clinical-fusion variants.

\subsection{Anatomy-specific visual evidence acquisition}

\begin{figure}[pos=htbp]
\centering
\includegraphics[width=0.95\textwidth]{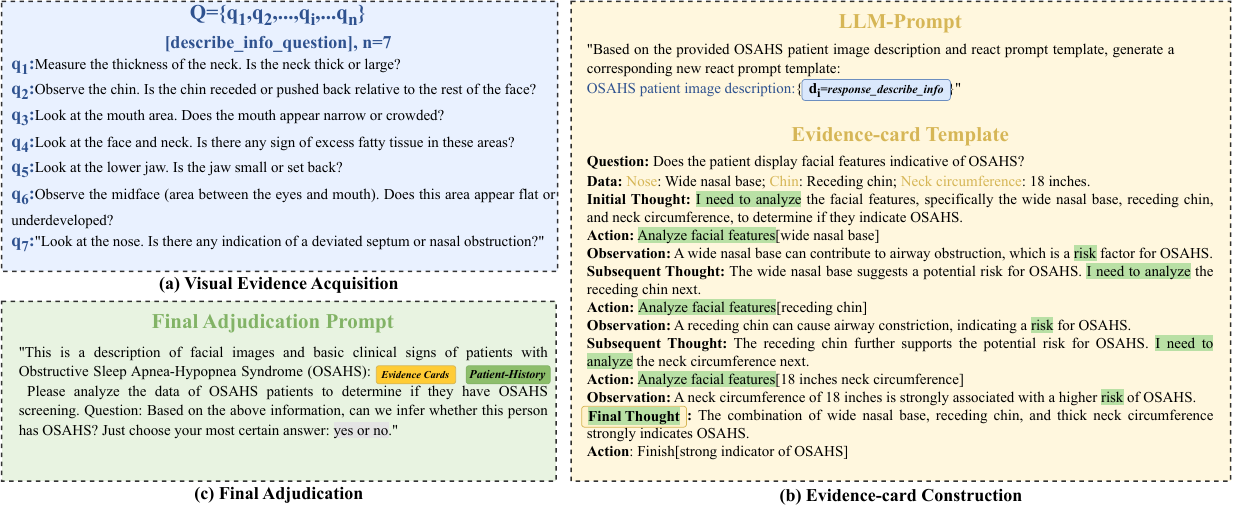}
\caption{\textbf{Prompt organization across the EviOSAHS workflow.}
The visual-observation prompt constrains the VLM to report anatomy-specific findings and visibility without making a clinical judgment. The evidence-card prompt converts each parsed visual observation into risk direction, evidence strength, confidence, and a concise evidence summary. The final adjudication prompt combines the evidence cards with a clean structured clinical summary to generate a balanced binary screening output.}
\label{fig:prompt_design}
\end{figure}

The visual stage constrains the VLM to act as an anatomy-specific observer rather than a direct clinical classifier. Instead of asking whether the subject has OSAHS, the frontal facial image is decomposed into $K=7$ fixed anatomical queries:
\begin{equation}
\mathcal{Q}
=
(q_i)_{i=1}^{7}
=
\bigl(
q_{\mathrm{neck}},
q_{\mathrm{chin}},
q_{\mathrm{mouth}},
q_{\mathrm{fat}},
q_{\mathrm{jaw}},
q_{\mathrm{midface}},
q_{\mathrm{nose}}
\bigr).
\label{eq:question_set}
\end{equation}
These queries target neck thickness, chin recession, oral crowding, face-and-neck fat accumulation, lower-jaw position, midface development, and visible nasal asymmetry or obstruction-related external cues. Table~\ref{tab:visual_evidence_schema} summarizes both the anatomical query inventory and the evidence-card schema. Figure~\ref{fig:prompt_design} illustrates how the prompt design constrains the model output to anatomical observation, visibility assessment, evidence-card construction, and final adjudication.

For each query $q_i$, the VLM receives only the facial image and the corresponding anatomical question. The raw observation and parsed visual tuple are generated as
\begin{equation}
o_{n,i}
=
f_{\mathrm{VLM}}
\bigl(
\mathcal{I}_n,q_i;T_{\mathrm{vis}}
\bigr),
\quad
u_{n,i}
=
p_{\mathrm{vis}}(o_{n,i})
=
\bigl(a_{n,i},t_{n,i},v_{n,i}\bigr),
\qquad i=1,\ldots,K .
\label{eq:vlm_observation}
\end{equation}
where $T_{\mathrm{vis}}$ is the visual-observation prompt template. The prompt explicitly prohibits OSAHS diagnosis, treatment recommendation, AHI inference, severity estimation, and use of clinical background information. The required output is restricted to three fields: anatomical target, short visual observation, and visibility level, with visibility categorized as high, medium, or uncertain.

Here, $a_{n,i}$ is the anatomical target, $t_{n,i}$ is the textual visual observation, and $v_{n,i}$ is the visibility level. The parsing step turns a free-form VLM response into a machine-checkable intermediate record. It also prevents the VLM from bypassing the evidence-acquisition stage and directly producing a clinical screening judgment. The visual stage can therefore be audited not only through its downstream contribution to the final label, but also through the parseability, visibility distribution, and target-level consistency of the observations it produces.

\subsection{Evidence-card construction}

The parsed visual observation $u_{n,i}$ is converted into an evidence card. This stage interprets the screening relevance of the observed anatomical cue without producing the final screening decision. Each card preserves the original anatomical target and visual observation while adding structured fields that describe whether the cue supports, argues against, or remains uncertain with respect to OSAHS screening risk:
\begin{equation}
e_{n,i}
=
f_{\mathrm{card}}
\bigl(
u_{n,i};T_{\mathrm{card}}
\bigr)
=
\bigl(
a_{n,i},
t_{n,i},
v_{n,i},
r_{n,i},
s_{n,i},
\kappa_{n,i},
m_{n,i}
\bigr),
\label{eq:evidence_card}
\end{equation}
where $T_{\mathrm{card}}$ is the evidence-card prompt template, $r_{n,i}$ is the risk direction, $s_{n,i}$ is the evidence strength, $\kappa_{n,i}$ is the model-reported confidence level, and $m_{n,i}$ is a concise evidence summary.

The categorical fields are deliberately constrained:
\begin{equation}
r_{n,i} \in \mathcal{R} = \{\mathrm{support}, \mathrm{against}, \mathrm{uncertain}\},
s_{n,i} \in \mathcal{S} = \{\mathrm{weak}, \mathrm{moderate}, \mathrm{strong}\},
\kappa_{n,i} \in \mathcal{K} = \{\mathrm{low}, \mathrm{medium}, \mathrm{high}\}.
\label{eq:card_categories}
\end{equation}

Visibility and confidence capture different aspects of uncertainty. Visibility describes whether the relevant anatomical region can be adequately inspected in the image, whereas confidence describes the model-reported certainty in the screening relevance assigned to the parsed observation. The confidence field is treated as a self-reported qualitative indicator rather than a calibrated probability. This controlled schema reduces free-form reasoning drift and makes intermediate evidence comparable across subjects and anatomical targets.

The evidence-card prompt asks the LLM to restate the visual observation, assess its screening relevance, assign categorical evidence fields, and summarize the evidence. It does not ask for a final diagnosis, severity category, or screening label. The required output fields, allowed categorical values, and workflow roles are specified in Table~\ref{tab:visual_evidence_schema} and Figure~\ref{fig:prompt_design}.

For each subject, the seven evidence cards form the ordered visual evidence list
\begin{equation}
\mathcal{E}_n
=
\bigl(e_{n,1},\ldots,e_{n,K}\bigr),
\quad K=7.
\label{eq:evidence_list}
\end{equation}
Compared with unconstrained chain-of-thought prompting~\cite{wei2022cot}, the evidence-card representation is more auditable: each intermediate statement is anchored to a specific anatomical target and assigned explicit direction, strength, confidence, and visibility.

\subsection{Final-only clinical adjudication}

The final stage integrates the visual evidence cards with the clean clinical summary. This is the only stage in which clinical information is introduced. We formalize the staged dependency as
\begin{equation}
\mathcal{E}_n
=
G_{\mathrm{vis}}
\bigl(
\mathcal{I}_n;\mathcal{Q},T_{\mathrm{vis}},T_{\mathrm{card}}
\bigr),
\quad
\hat{y}_n
=
f_{\mathrm{final}}
\bigl(
\mathcal{E}_n,\mathcal{C}_n;T_{\mathrm{final}}
\bigr).
\label{eq:final_only_dependency}
\end{equation}
The first term states that evidence cards are generated from the image and visual templates only. The second term states that the final screening prediction depends on both the evidence cards and the clean clinical summary. This dependency structure specifies precisely where clinical information enters the computational workflow.

Before final adjudication, evidence cards are organized into supporting, opposing, and uncertain groups:
\begin{equation}
\label{eq:evidence_groups}
\mathcal{E}^{+}_n = \{ e_{n, i} \mid r_{n, i} = \riskSupport \}, \quad \mathcal{E}^{-}_n = \{ e_{n, i} \mid r_{n, i} = \riskAgainst \}, \quad \mathcal{E}^{0}_n = \{ e_{n, i} \mid r_{n, i} = \riskUncertain \}.
\end{equation}
The corresponding counts are
\begin{equation}
\label{eq:evidence_counts}
\mathbf{N}_n = \bigl( N^{+}_n, N^{-}_n, N^{0}_n \bigr) = \bigl( |\mathcal{E}^{+}_n|, |\mathcal{E}^{-}_n|, |\mathcal{E}^{0}_n| \bigr).
\end{equation}
These counts are not used as a hard numerical threshold. Instead, they act as an explicit evidence-accounting mechanism: the final LLM must state how much supporting, opposing, and uncertain evidence is present before considering the clinical profile. This design makes the final adjudication process reviewable and allows discrepancies between evidence-card-derived counts and final reported counts to be checked.

The final adjudication prompt is constructed as
\begin{equation}
P^{\mathrm{final}}_n
=
\Format
\bigl(
\mathcal{E}_n,
\mathbf{N}_n,
\mathcal{C}_n,
T_{\mathrm{final}},
Q_{\mathrm{screen}}
\bigr),
\label{eq:final_prompt}
\end{equation}
where $T_{\mathrm{final}}$ is the balanced final-adjudication template and $Q_{\mathrm{screen}}$ asks for a binary screening-positive or screening-negative decision. Here, ``balanced'' refers to an instruction-level constraint requiring the final adjudicator to explicitly compare supporting, opposing, and uncertain evidence before issuing a label; it does not denote class rebalancing, probability calibration, or a numerical decision threshold.

The final output is constrained to include the reported evidence balance, a clinical risk level, a brief rationale, and the binary screening label:
\begin{equation}
h_n
=
f_{\mathrm{final}}
\bigl(
P^{\mathrm{final}}_n
\bigr)
=
\bigl(
\tilde{\mathbf{N}}_n,
\ell_n,
b_n,
\hat{y}_n
\bigr),
\label{eq:final_output}
\end{equation}
where $\tilde{\mathbf{N}}_n=(\tilde{N}^{+}_n,\tilde{N}^{-}_n,\tilde{N}^{0}_n)$ denotes the evidence balance reported by the final adjudicator, $\ell_n$ denotes the clinical risk level, $b_n$ denotes the brief final rationale, and $\hat{y}_n$ denotes the final binary screening label. The tildes indicate that these counts are reported in the final structured response and can be checked against the evidence-card-derived count vector $\mathbf{N}_n$.

Equivalently, the complete workflow can be summarized as
\begin{equation}
\label{eq:compact_composition}
\begin{aligned}
u_{n, i} &= p_{\mathrm{vis}} \bigl( f_{\mathrm{VLM}}(\mathcal{I}_n, q_i; T_{\mathrm{vis}}) \bigr), \quad e_{n, i} = f_{\mathrm{card}} \bigl( u_{n, i}; T_{\mathrm{card}} \bigr), \\
G_{\mathrm{vis}}(\mathcal{I}_n) &= \bigl( e_{n, i} \bigr)_{i=1}^{K}, \quad \hat{y}_n = f_{\mathrm{final}} \bigl( G_{\mathrm{vis}}(\mathcal{I}_n), g_{\mathrm{clin}}(\mathbf{x}_n); T_{\mathrm{final}} \bigr).
\end{aligned}
\end{equation}
Eq.~\ref{eq:compact_composition} highlights the central methodological distinction of EviOSAHS: the inner pathway is image-only and produces structured visual evidence, whereas the outer pathway introduces clinical information and performs final adjudication. The final answer is therefore not the result of a single unconstrained image-text prompt, but of staged visual evidence acquisition followed by evidence-grounded clinical adjudication.

For procedural clarity, Algorithm~\ref{alg:eviosahs_method} summarizes the inference workflow. It is included as an implementation-level summary and does not introduce additional modeling assumptions beyond the components defined above.

In this way, EviOSAHS treats OSAHS screening as an evidence acquisition and adjudication problem. The VLM serves as an anatomy-specific visual observer, while the LLM serves as the final adjudicator over explicit visual evidence and structured clinical context. Because the method returns both the final screening label and the evidence-card set, each output can be reviewed through its intermediate anatomical observations, evidence directions, evidence strengths, confidence levels, and final clinical rationale.

\begin{algorithm}[H]
\caption{EviOSAHS inference with image-only evidence acquisition and final-only clinical adjudication}
\label{alg:eviosahs_method}
\footnotesize
\begin{algorithmic}[1]
\Require Image $\mathcal{I}_n$; clinical profile $\mathbf{x}_n$; query set $\mathcal{Q}=(q_i)_{i=1}^{K}$; templates $\mathcal{T}=(T_{\mathrm{vis}},T_{\mathrm{card}},T_{\mathrm{final}},Q_{\mathrm{screen}})$
\Ensure Screening label $\hat{y}_n\in\{0,1\}$; evidence cards $\mathcal{E}_n$; structured final record $h_n$

\AlgStage{Phase I. Image-only evidence acquisition}
\State Initialize ordered evidence-card list: $\mathcal{E}_n \gets [\,]$

\For{$i=1,\ldots,K$}
    \State $o_{n,i} \gets f_{\mathrm{VLM}}(\mathcal{I}_n,q_i;T_{\mathrm{vis}})$ \Comment{clinical variables excluded}
    \State $u_{n,i} \gets p_{\mathrm{vis}}(o_{n,i})$ \Comment{anatomical target, observation, visibility}
    \State $e_{n,i} \gets f_{\mathrm{card}}(u_{n,i};T_{\mathrm{card}})$ \Comment{direction, strength, confidence}
    \State $\mathcal{E}_n \gets \mathcal{E}_n \mathbin{\|} [e_{n,i}]$
\EndFor

\AlgStage{Phase II. Clinical context reconstruction}
\State $\mathcal{C}_n \gets g_{\mathrm{clin}}(\mathbf{x}_n)$ \Comment{withheld until visual evidence is fixed}
\State Assert $\fields(\mathcal{C}_n)\cap\{\mathrm{AHI},z_n,y_n\}=\varnothing$

\AlgStage{Phase III. Evidence accounting and adjudication}
\State $(\mathcal{E}^{+}_n,\mathcal{E}^{-}_n,\mathcal{E}^{0}_n) \gets \GroupRisk(\mathcal{E}_n)$
\State $\mathbf{N}_n \gets (|\mathcal{E}^{+}_n|,|\mathcal{E}^{-}_n|,|\mathcal{E}^{0}_n|)$
\State $P^{\mathrm{final}}_n \gets \Format(\mathcal{E}_n,\mathbf{N}_n,\mathcal{C}_n,T_{\mathrm{final}},Q_{\mathrm{screen}})$
\State $h_n \gets f_{\mathrm{final}}(P^{\mathrm{final}}_n)$
\State $\hat{y}_n \gets \ExtractLabel(h_n)$ \Comment{rule-based parsing of constrained output}
\State \Return $(\hat{y}_n,\mathcal{E}_n,h_n)$
\end{algorithmic}
\end{algorithm}

\FloatBarrier

\section{Experiments and Results}

\subsection{Experimental design}

The experiments were designed to evaluate EviOSAHS as a high-sensitivity binary pre-polysomnography screening workflow for Obstructive Sleep Apnea-Hypopnea Syndrome (OSAHS). Following common reporting practice in medical image analysis studies, we first describe the cohort, label definition, implementation protocol, compared methods, evaluation metrics, and statistical analysis, and then present the main quantitative results, component ablations, visual-output audits, image-control experiments, subgroup behavior, error attribution, and representative evidence traces.

The experimental analysis addressed five questions:
(1) whether EviOSAHS improves binary screening performance over clinical-only prompting and direct multimodal prompting;
(2) whether the improvement is reflected at the paired subject level;
(3) which workflow components contribute most to the final operating point;
(4) whether the visual evidence stage produces structured and reviewable intermediate outputs and remains sensitive to image perturbation; and
(5) which subgroup patterns and error mechanisms characterize the current decision boundary.

All formal LLM/MLLM experiments used the full cohort for deterministic direct inference without cross-validation or fine-tuning on the target cohort. This setting reflects the intended use of EviOSAHS as a prompt-based multimodal screening workflow rather than a supervised classifier trained on the study cohort. Unless otherwise specified, all methods used the same binary label mapping, image set, structured clinical fields, deterministic decoding protocol, and output-parsing rule. Unknown, ambiguous, or unparsed outputs were counted as incorrect for all methods.

\begin{table}[pos=htbp]
\centering
\caption{\textbf{Cohort characteristics and binary label mapping.}
Continuous variables are reported as median and interquartile range unless otherwise indicated. For the primary binary screening endpoint, normal subjects were mapped to screening-negative and all mild, moderate, or severe OSAHS subjects were mapped to screening-positive.}
\label{tab:cohort}
\footnotesize
\setlength{\tabcolsep}{5.5pt}
\renewcommand{\arraystretch}{1.18}
\arrayrulecolor{tableRule}
\begin{tabularx}{\textwidth}{@{} P{3.8cm} P{4.9cm} Y @{}}
\toprule
\textbf{Variable} & \textbf{Result} & \textbf{Note} \\
\midrule
\TableBand{tableBlue}{3}{Cohort size and binary labels}
\addlinespace[1pt]
\rowcolor{tableStripe}
Total samples & 642 & Full cohort \\
Binary screening-positive & 584 & AHI-derived mild, moderate, or severe labels \\
\rowcolor{tableStripe}
Binary screening-negative & 58 & AHI-derived normal label \\
Original severity distribution & Normal 58 / mild 99 / moderate 95 / severe 390 & Used only to define binary screening labels \\
\addlinespace[3pt]
\TableBand{tableGreen}{3}{Demographic and anthropometric characteristics}
\addlinespace[1pt]
\rowcolor{tableStripe}
Sex & Male 546 / female 96 & -- \\
Age & 40.00 (33.00--49.00), range 4.00--82.00 & years \\
\rowcolor{tableStripe}
Neck circumference & 40.00 (37.00--41.00), range 26.00--51.00 & cm \\
BMI & 26.50 (24.40--29.67), range 15.60--44.60 & kg/m$^2$ \\
\rowcolor{tableStripe}
BMI category & Underweight 5 / healthy 197 / overweight 290 / obesity 144 / morbid obesity 6 & -- \\
Waist-hip ratio & 0.94 (0.90--0.98) & Two extreme values $>2.0$ were flagged during descriptive review \\
\addlinespace[3pt]
\TableBand{tableBlue}{3}{Comorbidities}
\addlinespace[1pt]
\rowcolor{tableStripe}
Hypertension & 99 (15.42\%) & -- \\
Diabetes & 14 (2.18\%) & -- \\
\rowcolor{tableStripe}
Heart disease & 2 (0.31\%) & -- \\
Hyperlipidemia & 15 (2.34\%) & -- \\
\bottomrule
\end{tabularx}
\end{table}

\subsection{Cohort and binary screening labels}

All experiments were conducted on a retrospective sleep-medicine cohort of 642 subjects evaluated for suspected sleep-related breathing disorders. The cohort was enriched for OSAHS-positive cases and should therefore be interpreted as a high-risk pre-polysomnography triage cohort rather than a general-population screening cohort. Each subject contained a frontal facial image, a structured clinical profile, and an AHI-derived reference severity category.

The original severity distribution included 58 normal, 99 mild, 95 moderate, and 390 severe cases. For the primary binary screening task, normal subjects were mapped to screening-negative and mild, moderate, or severe OSAHS subjects were mapped to screening-positive, yielding 58 screening-negative and 584 screening-positive subjects. The structured clinical profile included age, sex, BMI, neck circumference, waist-hip ratio, hypertension, diabetes, heart disease, and hyperlipidemia when available. AHI values, AHI-derived severity labels, and label-derived information were excluded from all model prompts to prevent leakage from the reference standard into the screening workflow. Table~\ref{tab:cohort} summarizes the cohort composition, binary label mapping, clinical variables, and comorbidity distribution used in the experiments.

\subsection{Implementation and inference protocol}

The primary EviOSAHS implementation used a Qwen-based VLM for anatomy-specific visual observation and a Qwen-based LLM for evidence-card construction and final adjudication. Direct multimodal baselines included InstructBLIP, LLaVA-1.6, and Qwen2.5-VL. The principal checkpoints were \texttt{Qwen2.5-VL-7B-Instruct}~\cite{bai2025qwen25vl}, \texttt{Qwen2.5-7B-Instruct}~\cite{yang2024qwen25}, \texttt{llava-v1.6-mistral-7b-hf}~\cite{liu2023llava}, and \texttt{Meta-Llama-3.1-8B-Instruct}~\cite{grattafiori2024llama3}. All experiments were implemented using the Hugging Face Transformers library.

Prompt templates, output schemas, and parsing rules were fixed before formal inference runs. Formal inference used deterministic decoding with temperature 0, top-$p=1.0$, and sampling disabled. Model outputs were parsed into binary screening-positive or screening-negative labels using rule-based parsing of the constrained answer format. Unknown, ambiguous, or unparsed outputs were counted as incorrect for all methods. All inference runs were conducted on an 80-GB-memory GPU.

\begin{table}[pos=htbp]
\centering
\caption{\textbf{Compared methods for binary OSAHS screening.}
All methods were evaluated on the same 642-subject cohort using the same binary label mapping, image set, structured clinical fields, deterministic decoding protocol, and output-parsing rule. Direct multimodal prompting methods received the facial image and clean clinical summary in a single prompt and were asked to produce a binary yes/no screening output. Two-stage and clinical-fusion variants were included to assess whether staged evidence acquisition, evidence-card organization, and final-only clinical adjudication contributed to screening performance. No method was fine-tuned on the target cohort.}
\label{tab:compared_methods}
\footnotesize
\setlength{\tabcolsep}{5pt}
\renewcommand{\arraystretch}{1.18}
\arrayrulecolor{tableRule}
\begin{tabularx}{\textwidth}{@{} P{3.1cm} P{5.3cm} Y @{}}
\toprule
\TableBand{tableBlue}{3}{Compared methods and workflow structure}
\midrule
\textbf{Method} & \textbf{Input and reasoning structure} & \textbf{Purpose} \\
\midrule
\rowcolor{tableStripe}
Clinical-only prompting 
& Clean structured clinical summary only 
& Text-only clinical reference \\

Direct InstructBLIP prompting 
& Image and clinical summary in a single yes/no prompt 
& Direct multimodal baseline \\

\rowcolor{tableStripe}
Direct LLaVA-1.6 prompting 
& Image and clinical summary in a single yes/no prompt 
& Strong non-Qwen direct multimodal baseline \\

Direct Qwen2.5-VL prompting 
& Image and clinical summary in a single yes/no prompt 
& Qwen direct multimodal baseline \\

\rowcolor{tableStripe}
Naive two-stage prompting 
& Visual-to-final workflow without full balanced evidence-card organization 
& Tests whether a simple two-stage design is sufficient \\

Early clinical-fusion variant 
& Clinical information introduced before final adjudication 
& Tests the effect of introducing clinical information before evidence finalization \\

\rowcolor{tableStripe}
Single-model Qwen variant 
& Single-model visual-to-reason-to-final workflow 
& Tests whether one model can perform all stages \\

\textbf{EviOSAHS} 
& Seven-question visual decomposition, structured evidence cards, evidence strength, and final-only clinical adjudication 
& Primary proposed method \\
\bottomrule
\end{tabularx}
\end{table}

\subsection{Compared methods}

Table~\ref{tab:compared_methods} summarizes the compared methods. Methods are reported by input modality and reasoning structure rather than by internal experiment identifiers.

For component analysis, we evaluated five EviOSAHS ablations: removing session-level reasoning organization, removing the clean structured clinical summary, removing evidence strength, removing balanced final adjudication, and replacing seven-question visual decomposition with single-pass visual extraction. Each ablation modified one component while keeping the cohort, label mapping, decoding settings, and parsing rule unchanged.

\subsection{Evaluation metrics and statistical analysis}

Because the intended use case is high-sensitivity pre-PSG triage, sensitivity and false-negative rate (FNR) were treated as safety-oriented metrics. Accuracy was reported as the overall binary screening metric, and F1-score was used to summarize classification performance under class imbalance. The tendency toward unnecessary positive predictions was analyzed through prediction distribution, error composition, and representative failure patterns rather than being used as the primary operating-point selection criterion.

Let TP, TN, FP, and FN denote true positives, true negatives, false positives, and false negatives. The binary metrics were computed as:
\begin{equation}
\begin{alignedat}{2}
\operatorname{Acc}
&=
\frac{\mathrm{TP}+\mathrm{TN}}
{\mathrm{TP}+\mathrm{TN}+\mathrm{FP}+\mathrm{FN}},
\qquad&
\operatorname{Sens}
&=
\frac{\mathrm{TP}}
{\mathrm{TP}+\mathrm{FN}},
\\[0.8ex]
\mathrm{F}_{1}
&=
\frac{2\mathrm{TP}}
{2\mathrm{TP}+\mathrm{FP}+\mathrm{FN}},
\qquad&
\operatorname{FNR}
&=
\frac{\mathrm{FN}}
{\mathrm{TP}+\mathrm{FN}} .
\end{alignedat}
\label{eq:eval_metrics}
\end{equation}

Paired method comparisons used exact McNemar tests on sample-level correctness. For each paired comparison, discordant counts corresponded to subjects correctly classified by EviOSAHS but incorrectly classified by the comparator, and subjects correctly classified by the comparator but incorrectly classified by EviOSAHS.

\begin{table}[pos=htbp]
\caption{\textbf{Main binary OSAHS screening performance on the 642-subject cohort.}
Values are reported as percentages. Bold indicates the best performance; for FNR, lower is better. 
Unknown, ambiguous, or unparsed outputs were counted as incorrect. 
Unnecessary positive predictions and prediction distribution are summarized separately in Figure~\ref{fig:main_behavior}.}
\label{tab:main_results}
\centering
\normalsize
\setlength{\tabcolsep}{8pt}
\renewcommand{\arraystretch}{1.30}
\arrayrulecolor{tableRule}

\begin{tabularx}{\textwidth}{@{} Y
    S[table-format=3.2]
    S[table-format=3.2]
    S[table-format=3.2]
    S[table-format=3.2] @{}}
\toprule
\TableBandMain{tableBlue}{5}{Main binary OSAHS screening performance}
\midrule
\textbf{Method} 
& {\textbf{Accuracy (\%)}} 
& {\textbf{Sensitivity (\%)}} 
& {\textbf{F1-score (\%)}} 
& {\textbf{FNR (\%)}} \\
\midrule

\rowcolor{tableStripe}
Clinical-only prompting
& 74.61
& 75.34
& 84.37
& 24.66 \\

Direct InstructBLIP prompting
& 9.03
& 0.00
& 0.00
& 100.00 \\

\rowcolor{tableStripe}
Direct LLaVA-1.6 prompting
& 73.83
& 74.66
& 83.85
& 25.34 \\

Direct Qwen2.5-VL prompting
& 9.97
& 1.54
& 3.02
& 98.46 \\
\addlinespace[2pt]

\rowcolor{tableStripe}
Naive two-stage prompting
& 64.49
& 64.04
& 76.64
& 35.96 \\
\addlinespace[2pt]

Early clinical-fusion variant
& 85.67
& 91.61
& 92.08
& 8.39 \\

\rowcolor{tableStripe}
Single-model Qwen variant
& 85.36
& 89.55
& 91.75
& 10.45 \\

\textbf{EviOSAHS}
& {\bfseries 88.47}
& {\bfseries 94.86}
& {\bfseries 93.74}
& {\bfseries 5.14} \\

\bottomrule
\end{tabularx}
\end{table}

\subsection{Experimental results}

\subsubsection{Main binary screening performance}

\begin{figure}[pos=h]
\centering
\includegraphics[width=0.90\textwidth]{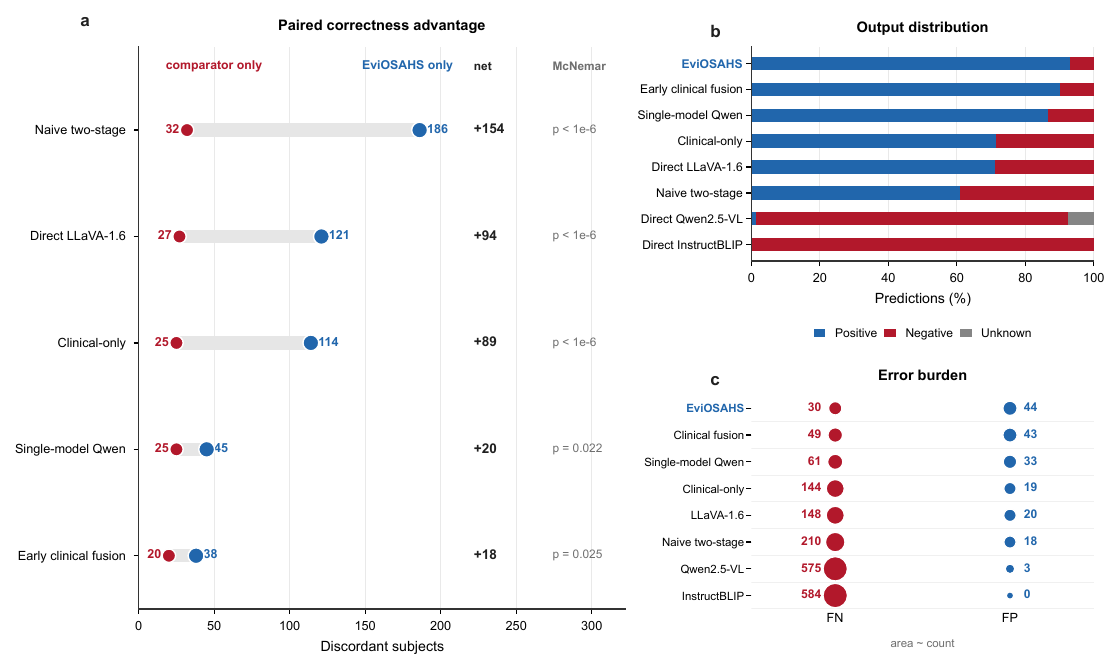}
\caption{\textbf{Main prediction behavior and paired sample-level comparison.}
(A) Prediction distribution across methods, showing screening-positive, screening-negative, and unknown outputs. This panel displays operating-point behavior without duplicating the percentage metrics in Table~\ref{tab:main_results}.
(B) Error composition across methods, showing false-negative and unnecessary positive prediction counts. This panel visualizes missed-case reduction and unnecessary positive predictions as raw counts.
(C) Paired discordance between EviOSAHS and selected comparator methods. Bars extending to the right indicate subjects correctly classified by EviOSAHS but incorrectly classified by the comparator; bars extending to the left indicate subjects correctly classified by the comparator but incorrectly classified by EviOSAHS. Exact McNemar $p$ values are annotated for each paired comparison.}
\label{fig:main_behavior}
\end{figure}

Table~\ref{tab:main_results} reports the main binary screening performance. EviOSAHS achieved the most favorable high-sensitivity operating point among the compared methods, with the highest accuracy, sensitivity, and F1-score, together with the lowest FNR. Relative to the clinical-only baseline and the strongest direct multimodal baseline, EviOSAHS reduced false negatives from more than 140 cases to 30 cases, indicating improved missed-case avoidance for pre-PSG triage. This gain was accompanied by a positive-shifted decision boundary, consistent with a screening-oriented workflow rather than a definitive diagnostic rule-out model.

Direct multimodal prompting showed heterogeneous and unstable behavior. Direct InstructBLIP prompting collapsed to all-negative outputs; direct Qwen2.5-VL prompting produced mostly negative or unknown outputs; and direct LLaVA-1.6 prompting remained close to the clinical-only baseline. The early clinical-fusion and single-model variants approached EviOSAHS more closely than the direct baselines, but both retained higher FNRs. These results indicate that the performance gain was not simply due to using a stronger multimodal backbone, but to the structured evidence acquisition and final adjudication workflow.

Figure~\ref{fig:main_behavior} further characterizes this operating point through prediction distributions, error counts, and paired sample-level discordance, complementing the aggregate percentages in Table~\ref{tab:main_results}.

\begin{table}[pos=h]
\centering
\caption{\textbf{Component ablation of EviOSAHS on the 642-subject cohort.}
Each row removes or modifies one component of the full EviOSAHS workflow while keeping the same cohort, binary label mapping, deterministic decoding protocol, and output-parsing rule. Values are reported as percentages unless otherwise indicated. $\Delta$ Sens. and $\Delta$ Acc. denote absolute percentage-point changes relative to full EviOSAHS. McNemar $p$ values compare each ablation with full EviOSAHS at the paired sample level. FNR = false-negative rate.}
\label{tab:ablation}
\footnotesize
\setlength{\tabcolsep}{3pt}
\renewcommand{\arraystretch}{1.20}
\arrayrulecolor{tableRule}
\begin{tabularx}{\textwidth}{@{} Y
    S[table-format=2.2]
    S[table-format=2.2]
    S[table-format=2.2]
    S[table-format=2.2]
    S[table-format=+2.2]
    S[table-format=+2.2]
    Z{1.35cm} @{}}
\toprule
\TableBand{tableGreen}{8}{Component ablation of the EviOSAHS workflow}
\midrule
\textbf{Configuration} 
& {\textbf{Accuracy}} 
& {\textbf{Sensitivity}} 
& {\textbf{F1-score}} 
& {\textbf{FNR}} 
& {\textbf{$\Delta$ Acc.}} 
& {\textbf{$\Delta$ Sens.}} 
& \makecell{\textbf{McNemar}\\\textbf{$p$}} \\
\midrule

\textbf{Full EviOSAHS}
& {\bfseries 88.47}
& {\bfseries 94.86}
& {\bfseries 93.74}
& {\bfseries 5.14}
& \multicolumn{1}{c}{--}
& \multicolumn{1}{c}{--}
& -- \\
\addlinespace[2pt]

w/o clean structured clinical summary
& 86.29
& 93.15
& 92.52
& 6.85
& -2.18
& -1.71
& 0.0436 \\

\rowcolor{tableStripe}
w/o evidence strength
& 84.42
& 88.70
& 91.20
& 11.30
& -4.05
& -6.16
& 0.00219 \\

w/o balanced final adjudication
& 73.99
& 75.68
& 84.11
& 24.32
& -14.49
& -19.18
& $<0.001$ \\

\rowcolor{tableStripe}
w/o seven-question visual decomposition
& 71.65
& 71.92
& 82.19
& 28.08
& -16.82
& -22.95
& $<0.001$ \\

\bottomrule
\end{tabularx}
\end{table}

\subsubsection{Ablation}

Table~\ref{tab:ablation} reports component ablation results. The largest degradations occurred when the two core structural constraints were removed. Replacing seven-question visual decomposition with single-pass visual extraction reduced sensitivity by 22.95 percentage points and accuracy by 16.82 percentage points, indicating that anatomy-specific evidence acquisition was critical for missed-case reduction. Removing balanced final adjudication reduced sensitivity by 19.18 percentage points and accuracy by 14.49 percentage points, showing that explicit comparison of supporting, opposing, and uncertain evidence was necessary for stable high-sensitivity behavior.

Removing evidence strength caused a smaller but clinically relevant drop, increasing FNR from 5.14\% to 11.30\%. Removing the clean structured clinical summary produced a modest decrease in both sensitivity and accuracy, supporting the contribution of standardized clinical context at the final stage. In contrast, removing session-level reasoning organization preserved overall accuracy, suggesting that the principal gain came from structured evidence acquisition and constrained adjudication rather than from free-form reasoning traces alone.

\begin{figure}[pos=h]
\centering
\includegraphics[width=0.90\textwidth]{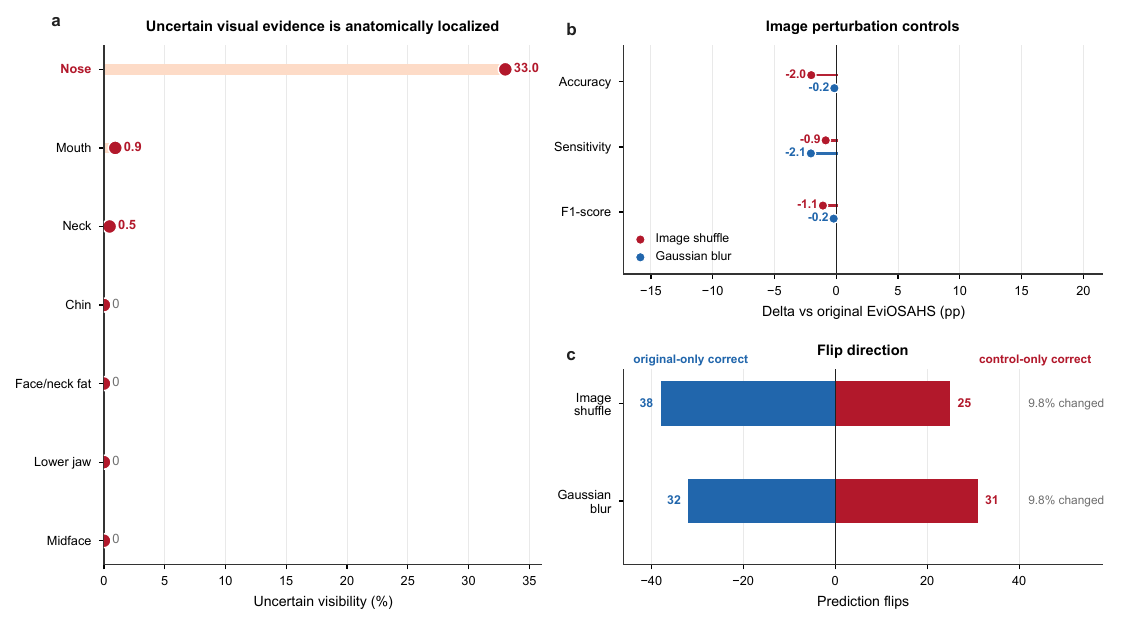}
\caption{\textbf{Visual-output audit and image-control analysis.}
(A) Visibility distribution across the seven anatomy-specific visual questions. Each bar summarizes the proportion of outputs categorized as high, medium, or uncertain visibility for a given anatomical target. The audit was conducted across $7 \times 642 = 4{,}494$ anatomical question sessions.
(B) Metric changes under image-shuffle and Gaussian-blur controls relative to unperturbed EviOSAHS. Bars show absolute percentage-point changes in accuracy, sensitivity, and F1-score.
(C) Prediction flip counts induced by each image-control condition, defined as the number of subjects whose binary prediction differed from the unperturbed EviOSAHS output.}
\label{fig:visual_control}
\end{figure}

\subsubsection{Visual-output audit and image controls}

Figure~\ref{fig:visual_control} summarizes visual-output audit and image-control experiments. Across $7 \times 642 = 4{,}494$ anatomy-specific visual question sessions, structured parsing succeeded for all outputs, and the overall irrelevant-content rate was 0.07\%. High visibility exceeded 96\% for all anatomical targets except the nose target, which showed the largest uncertain-visibility fraction. These results support the formatting reliability and reviewability of the image-only evidence acquisition stage, but they should not be interpreted as expert-annotated anatomical accuracy.

Image-shuffle and Gaussian-blur controls produced small aggregate changes in accuracy and sensitivity but induced prediction flips in 63 of 642 subjects for each perturbation. These controls indicate that individual decisions can be sensitive to image perturbation even when aggregate metrics remain relatively stable. Therefore, the image-control experiments were used to characterize prediction variability under visual perturbation, not to claim robustness to all possible image transformations.

\begin{figure}[pos=ht]
\centering
\includegraphics[width=0.90\textwidth]{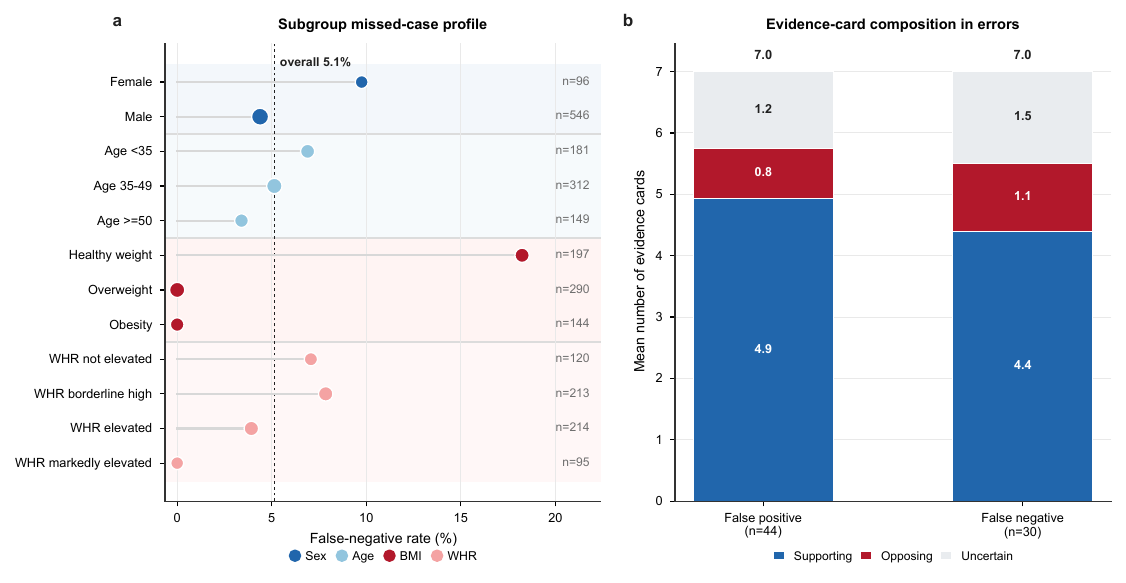}
\caption{\textbf{Subgroup behavior and error attribution.}
(A) False-negative rates of EviOSAHS across selected demographic and clinical strata available in the subgroup analysis, including sex, age group, BMI category, and waist-hip-ratio category. Severity-stratified results are not shown because severity grading is not the primary endpoint of this study.
(B) Mean evidence-card counts in false-positive and false-negative cases. Bars show the average number of supporting, opposing, and uncertain visual evidence cards per case. This panel summarizes error-case evidence patterns rather than calibrated causal explanations.}
\label{fig:subgroup_error}
\end{figure}

\subsubsection{Subgroup behavior and error attribution}

Figure~\ref{fig:subgroup_error} reports descriptive subgroup behavior and evidence-level error attribution. Subgroup FNR differed across demographic and clinical strata. Higher missed-case rates were observed in lower-risk phenotypic groups, including healthy-weight subjects, whereas overweight and obesity groups showed lower FNRs. Sex- and age-stratified results showed additional heterogeneity. WHR-stratified results should be interpreted cautiously because WHR included flagged extreme values in the descriptive cohort review.

False-positive and false-negative cases both contained multiple supporting visual cues on average. False positives had a mean of 4.93 supporting, 0.82 opposing, and 1.25 uncertain evidence cards, whereas false negatives had a mean of 4.40 supporting, 1.10 opposing, and 1.50 uncertain evidence cards. This pattern suggests that the current evidence-strength scheme can accumulate weak supporting cues into positive decisions, contributing to unnecessary positive predictions. These analyses are descriptive and are intended to localize failure modes rather than establish causal explanations.

\begin{figure}[pos=h]
\centering
\includegraphics[width=0.98\textwidth]{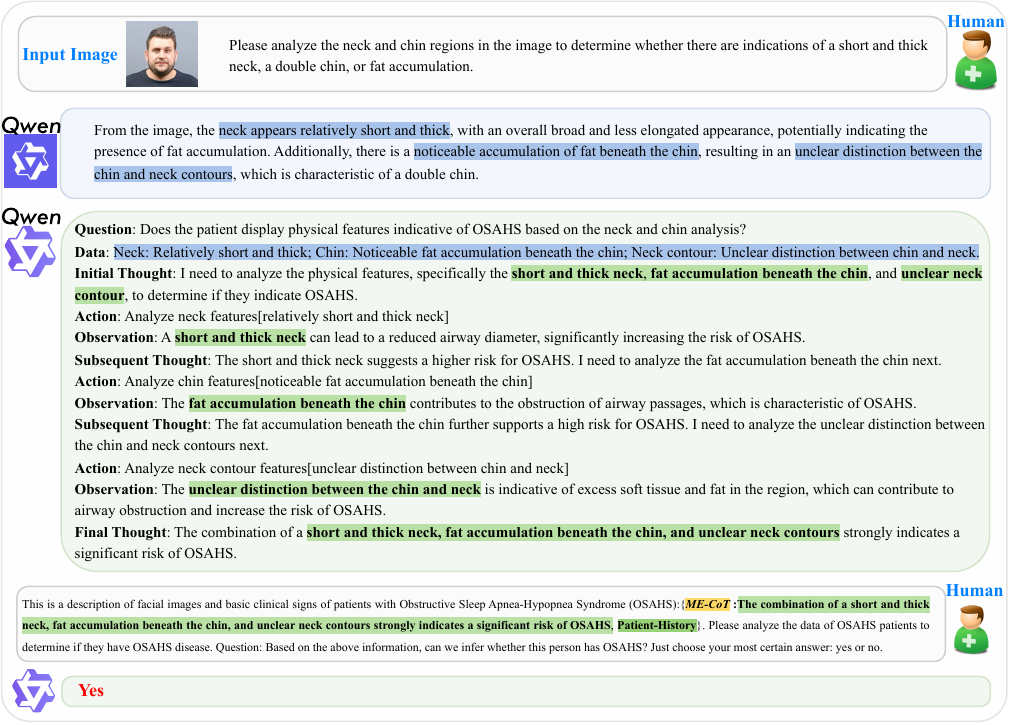}
\caption{\textbf{Representative EviOSAHS evidence trace.}
The example illustrates how a final screening output can be traced back to image-derived anatomical observations, evidence-card assignments, clinical context, final rationale, and comparator predictions. The case is illustrative and was not used as quantitative evidence of performance. Additional representative operating patterns are summarized in Table~\ref{tab:case_trace_summary}.}
\label{fig:evidence_traces}
\end{figure}

\begin{table}[pos=h]
\centering
\caption{\textbf{Representative case-study trace summary.}
Each row summarizes one full trace from the case-study evidence file. Evidence counts report the number of structured evidence cards assigned as supporting, opposing, or uncertain by EviOSAHS. The EviOSAHS output is separated and shaded to distinguish the proposed method from comparator outputs. Comparator outputs are shown in the fixed order clinical-only prompting, direct LLaVA-1.6 prompting, naive two-stage prompting, single-pass visual extraction, and single-model balanced prompting. Green indicates a correct case-level prediction and red indicates an incorrect case-level prediction. These cases were selected to illustrate reviewability and typical operating patterns, not to estimate performance.}
\label{tab:case_trace_summary}
\scriptsize
\setlength{\tabcolsep}{2.2pt}
\renewcommand{\arraystretch}{1.16}
\arrayrulecolor{tableRule}
\begin{tabularx}{\textwidth}{@{} P{2.05cm}
    Z{0.7cm}
    Z{0.8cm}
    Z{1.05cm}
    Z{1.05cm}
    Z{1.0cm}
    Z{1.55cm}
    Y @{}}
\toprule
\TableBand{tableGreen}{8}{Representative evidence-trace cases}
\midrule
\textbf{Case pattern}
& \makecell{\textbf{Case}\\\textbf{ID}}
& \textbf{Gold}
& \makecell{\textbf{EviOSAHS}\\\textbf{output}}
& \makecell{\textbf{Cards}\\\textbf{S/A/U}}
& \makecell{\textbf{Clinical}\\\textbf{risk}}
& \makecell{\textbf{Comparator}\\\textbf{outputs}}
& \textbf{Trace interpretation} \\
\midrule

\rowcolor{tableStripe}
True-positive recovery
& 626
& Yes
& \cellcolor{tableGreen}\textbf{Yes}
& 5/1/1
& Medium
& \makecell{\textcolor{red!70!black}{No/No/No}\\\textcolor{red!70!black}{No/No}}
& \textbf{EviOSAHS-only recovery:} multiple weak anatomical supports, including chin, mouth, lower-jaw, midface, and nasal cues, shifted the final adjudication to screening-positive despite low-risk clinical variables and uniformly negative comparator outputs. \\

Structure-driven contribution
& 10
& Yes
& \cellcolor{tableGreen}\textbf{Yes}
& 6/0/1
& Medium
& \makecell{\textcolor{red!70!black}{No/No/No}\\\textcolor{red!70!black}{No/No}}
& \textbf{Structure-driven recovery:} seven-question decomposition accumulated several localized weak supports that were missed by the clinical-only and less structured comparator workflows. \\

\rowcolor{tableStripe}
calibration limitation
& 50
& No
& \cellcolor{red!50}\textbf{Yes}
& 6/0/1
& Medium
& \makecell{\textcolor{green!45!black}{No/No/No}\\\textcolor{green!45!black}{No}/\textcolor{red!70!black}{Yes}}
& \textbf{unnecessary positive predictions:} multiple weak supporting visual cues can outweigh a clinically low-risk profile and produce unnecessary referral. \\

True negative
& 486
& No
& \cellcolor{tableGreen}\textbf{No}
& 4/1/2
& Low
& \makecell{\textcolor{green!45!black}{No/No/No}\\\textcolor{green!45!black}{No/No}}
& \textbf{Concordant true negative:} mixed weak visual evidence was not sufficient to override a low-risk clinical profile, resulting in a screening-negative decision shared by all comparators. \\

\rowcolor{tableStripe}
False-negative limitation
& 73
& Yes
& \cellcolor{red!50}\textbf{No}
& 5/1/1
& Low
& \makecell{\textcolor{red!70!black}{No/No/No}\\\textcolor{red!70!black}{No}/\textcolor{green!45!black}{Yes}}
& \textbf{False-negative boundary:} despite several weak supporting cues, the final adjudicator assigned low clinical risk and produced a screening-negative output, showing the residual missed-case risk of the workflow. \\

\bottomrule
\end{tabularx}
\begin{tablenotes}[flushleft]
\footnotesize
\item Abbreviations: S/A/U, supporting/against/uncertain evidence-card counts. Comparator outputs are binary screening predictions in the order listed in the caption. Green marks correct case-level predictions; red marks incorrect predictions. The full start-to-end prompts, raw responses, evidence cards, final responses, and comparator traces are retained in the case-study evidence file.
\end{tablenotes}
\end{table}

\subsubsection{Representative evidence traces}

Figure~\ref{fig:evidence_traces} and Table~\ref{tab:case_trace_summary} provide a qualitative audit of individual EviOSAHS decisions. The purpose of this analysis is not to add another performance metric, but to show how a final screening output can be traced back to the image-derived observations, structured evidence-card directions, clinical context, final adjudication, and comparator behavior. This trace-level view is important because the quantitative results establish that EviOSAHS improves missed-case reduction, whereas the case studies show how those decisions are formed and where they may fail.

The complete trace in Figure~\ref{fig:evidence_traces} illustrates a screening-positive subject for whom all selected comparator workflows produced screening-negative outputs, while EviOSAHS produced the correct screening-positive decision. The structured clinical profile alone suggested low baseline risk because the subject had healthy BMI and no recorded cardiometabolic comorbidity. In contrast, the anatomy-specific decomposition produced five weak supporting evidence cards, one opposing card, and one uncertain card. The final adjudicator therefore did not rely on a single decisive visual cue; instead, it converted the accumulation of mild craniofacial findings into a medium-risk final assessment. This example illustrates the intended role of EviOSAHS: the positive decision is recoverable from explicit intermediate evidence rather than from an opaque direct yes/no generation.

Table~\ref{tab:case_trace_summary} extends this audit to five representative cases and visually separates the EviOSAHS output from the comparator outputs. Green cells or text indicate correct case-level predictions, whereas red cells or text indicate incorrect predictions. This formatting makes the case-level advantage of EviOSAHS explicit in the first two screening-positive examples: both had clinically low-risk profiles and were missed by clinical-only prompting, direct LLaVA-1.6 prompting, naive two-stage prompting, single-pass visual extraction, and the single-model balanced variant, but EviOSAHS recovered them by accumulating localized weak supports across the chin, mouth, jaw, and midface evidence cards. These cases help explain why the seven-question decomposition improved sensitivity in the ablation study: it made mild but repeated anatomical cues available to the final adjudicator instead of compressing them into a single direct multimodal response.

The limitation cases show the other side of the same operating point. In the false-positive case, a reference-negative subject received six supporting and one uncertain evidence cards, leading EviOSAHS to issue a screening-positive output despite a clinically low-risk profile. This pattern is consistent with the error analysis in Figure~\ref{fig:subgroup_error}: when several weak supporting visual cues accumulate, the high-sensitivity adjudication strategy can favor referral even when the reference label is negative. Conversely, the true-negative case shows that weak visual supports do not automatically force a positive decision; when the clinical profile remained low risk and the evidence was mixed, EviOSAHS retained a screening-negative output shared by all comparators.

The false-negative case identifies a residual missed-case mechanism. Although the evidence-card summary contained several weak supporting cues, the final adjudicator assigned low clinical risk and produced a screening-negative decision, whereas the single-model balanced comparator produced a screening-positive output. This case suggests that the final aggregation stage can still underweight weak visual evidence when the structured clinical profile appears reassuring. Overall, the case studies support two conclusions: EviOSAHS improves reviewability by preserving inspectable intermediate evidence, but further calibration is needed to control how multiple weak visual cues are aggregated into referral-oriented decisions. These examples are illustrative only; formal quantitative comparisons are reported in Table~\ref{tab:main_results}, Table~\ref{tab:ablation}, and Figures~\ref{fig:main_behavior}--\ref{fig:subgroup_error}.


\section{Limitations and Future Directions}

The findings should be interpreted within the context of the study design and intended use. The cohort was retrospectively collected from a single sleep-medicine setting and was enriched for OSAHS-positive subjects. Therefore, the reported performance reflects a high-risk pre-polysomnography triage scenario rather than general-population screening. Broader validation across institutions, acquisition environments, demographic groups, disease-prevalence settings, and clinical workflows will be necessary to determine whether the observed operating behavior is transportable to more diverse screening settings.

EviOSAHS is intended to support high-sensitivity screening rather than to provide definitive diagnosis or severity grading. Although the framework reduced missed screening-positive cases, the current decision behavior may still lead to unnecessary referrals when multiple weak risk cues accumulate. The output should therefore be interpreted as triage support for clinician review and confirmatory PSG, not as an autonomous medical decision. Future work should investigate calibrated operating-point control, referral-threshold adjustment, uncertainty-aware outputs, and clinician-in-the-loop triage strategies to better balance missed-case reduction with referral burden.

The visual-output audit focused on structural parseability, visibility, irrelevant-content rate, and evidence-card consistency. These analyses support the reviewability of the image-only evidence acquisition stage, but they do not establish expert-validated anatomical accuracy. The intermediate visual evidence should therefore be interpreted as model-generated, reviewable anatomical evidence rather than clinician-confirmed craniofacial assessment. Future studies should incorporate expert craniofacial annotations and richer anatomical references, including multi-view facial images, lateral cephalometry, three-dimensional facial scans, or complementary physiological signals, to clarify which visual cues are clinically reliable.

The framework also depends on the behavior of the underlying foundation models and prompt implementation. Although the staged evidence-card design improves auditability, model-family dependence, prompt sensitivity, output variability, uncertainty estimation, privacy protection, and deployment safety remain unresolved. Prospective workflow studies are needed to determine whether evidence-grounded multimodal screening can improve real-world pre-PSG referral decisions, clinician efficiency, and patient-level outcomes.

\section{Conclusion}

This study presents EviOSAHS, an evidence-grounded multimodal workflow for high-sensitivity binary OSAHS screening from frontal facial images and structured clinical profiles. The framework separates image-only anatomical evidence acquisition from final clinical adjudication, converting localized visual observations into structured evidence cards before integrating them with standardized clinical context. In a high-risk sleep-medicine cohort, EviOSAHS reduced missed screening-positive cases compared with clinical-only prompting, direct multimodal prompting, and naive two-stage variants while preserving a reviewable evidence trace. These findings suggest that structured visual evidence decomposition can make general-purpose multimodal foundation models more suitable for clinician-reviewable pre-PSG triage. EviOSAHS should be regarded as a screening-support tool rather than a diagnostic or severity-grading system, and prospective multicenter validation with calibrated operating-point control is required before clinical deployment.


\section*{Acknowledgments}
This work is supported by Shanghai Municipal Natural Science Foundation (23ZR1425400).

\section*{Ethics approval and consent to participate}
Ethics approval was obtained from the Ethics Com-
mittee of the EENT Hospital of Fudan University (No.2022140)
and the study was registered in the Chinese Clinical Trial Registry
(ChiCTR2300069223). Informed consent was obtained from all
patients before the procedure.

\section*{Data Availability}

The clinical data used in this study contain sensitive patient information, including facial images and structured clinical records, and therefore cannot be made publicly available due to privacy, ethical, and institutional restrictions. De-identified tabular data and derived evaluation labels may be made available from the corresponding author upon reasonable request, subject to institutional approval, ethical review, and a data use agreement. Raw facial images and identifiable clinical records cannot be shared publicly.

\section*{Code Availability}
The source code for implementing EviOSAHS is publicly available at \textit{https://github.com/Leonard-zc/EviOSAHS}. The repository includes the fixed prompt templates, evidence-card schemas, inference scripts, output-parsing rules, evaluation scripts, and statistical-analysis code used in this study. The repository does not contain raw facial images, identifiable clinical records, AHI values, or any other sensitive patient-level data. 

\subsection*{Conflict of interest}

The authors declare that they have no competing interests.


\bibliographystyle{unsrt}

\bibliography{mybibfile}



\end{document}